\documentclass{amsart}
\usepackage[nocompress]{cite}
\usepackage[pdftex]{graphicx}
\usepackage{amsmath}
\usepackage{fixltx2e}
\usepackage{color}
\usepackage{booktabs}
\usepackage{multirow}

\graphicspath{{fig/}}

\newcommand{\R}{\mathbf{R}}

\newcommand{\x}{\boldsymbol{x}}

\renewcommand{\u}{\boldsymbol{u}}

\newcommand{\p}{\boldsymbol{p}}

\newcommand{\q}{\boldsymbol{q}}
\newcommand{\z}{\boldsymbol{z}}

\renewcommand{\b}{\boldsymbol{b}}

\newcommand{\T}{^\mathrm{T}}

\newcommand{\foetwo}{2${}\times{}$2}
\newcommand{\foethree}{3${}\times{}$3}
\newcommand{\foefive}{5${}\times{}$5}

\definecolor{dkmag}{rgb}{0.5,0,0.5}

\begin{document}
\title{Continuous Optimization for Fields of Experts Denoising Works}

\author{Petter Strandmark}
\address[Petter Strandmark]{Google Inc., Lund University}
\email{petter@maths.lth.se}
\author{Sameer Agarwal}
\address[Sameer Agarwal]{Google Inc.}
\email[Corresponding author]{sameeragarwal@google.com}

\begin{abstract}
Several recent papers use image denoising with a Fields of Experts prior to benchmark discrete optimization methods~\cite{ishikawa2009hocr,ishikawa2009,kahl2012,ishikawa2011,fix2011,lan2006,potetz2007}. 
We show that a non-linear least squares solver significantly outperforms all known discrete methods on this problem.
\end{abstract}

\maketitle

\section{Introduction}
For many optimization problems, a local model built using derivatives simply does not give any useful information about the global structure, making it hard for continuous methods to find good solutions. For some of these problems, the discrete optimization methods commonly used in computer vision and image ana\-lysis are able to avoid undesired local minima; see~\cite{boykov2001,crandall2011discrete}. 

Fields of Experts (FoE) is a sophisticated prior on the statistics of natural images\cite{foe}. It has a larger clique structure that is capable of capturing higher order interactions around each image pixel than models based on pairwise interactions. One area where the FoE priors have been used to great success is image denoising. For example, the recent state-of-the-art results in image denoising~\cite{nowozin2012} use FoE as one part in a more complicated machine learning system.

There seems to be a general sense that MAP inference in problems arising from the use of FoE models is hard and that continuous optimization methods may not be suitable for it. Thus an increasingly sophisticated (and expensive) array of discrete optimization methods have been developed to solve them 
~\cite{ishikawa2009hocr,ishikawa2009,kahl2012,ishikawa2011,fix2011,lan2006,potetz2007}. In this short paper we argue that a simple continuous optimization method can be used to solve the FoE denoising problem cheaply and effectively. We provide an open-source implementation of this method.

\section{Denoising using Fields of Experts}
Given a noisy image $\u$, the negative log-likelihood for an image $\x$ using the FoE prior is
\begin{equation}
  \label{eq:problem}
  f(\x) = \sum_{i=1}^n \frac{(x_i - u_i)^2}{2\sigma^2} 
        + \sum_{P\in \mathcal{P}} \,\,\,\, \sum_{k=1}^K \alpha_k \log\left( 1 + 
            \frac{1}{2}\left( \b_k\T\x_P \right)^2 \right),
\end{equation}
where $P$ is an image patch in the set of all $m\times m$ patches $\mathcal{P}$ of $\x$ and $\sigma$ is the standard deviation of the Gaussian noise in $\u$.
The coefficients $\alpha_k$ and the filters $\b_k$ for $k=1,\ldots,K$ are estimated from a database of natural images\cite{foe}. This, the problem of denoising $\u$ can be formulated as the finding image $\x$ that minimizes~\eqref{eq:problem}.

\section{Fusion Moves}
Fusion move is a common discrete optimization method used in image analysis~\cite{lempitsky2010fusion}. 
 Let $\p,\q\in \R^n$ be two candidate solutions to a minimization problem $\min_{\x\in\R^n} f(\x)$. A new solution can be formed by picking (``fusing'') components from $\p$ and $\q$ independently according to an indicator vector $\z$:
\begin{equation}
 \min_{\z\in\{0,1\}^n} f\Big((1-\z)\cdot \p + \z\cdot \q\Big).
\end{equation}

Computing the optimal $\z$ is a discrete optimization problem which can be solved using roof duality~\cite{lempitsky2010fusion,kahl2012}. This framework includes $\alpha$-expansion~\cite{boykov2001} as a special case.

 All of the fastest reported methods for minimizing~\eqref{eq:problem} are based on fusion moves \cite{ishikawa2009hocr,ishikawa2009,ishikawa2011,fix2011}. Generating good candidates is crucial and can be done in many ways. Two types of candidates have been used for FoE denoising: (i) blurring and randomly perturbing a current solution \cite{ishikawa2009hocr,ishikawa2011} and (ii) generating a candidate from the gradient of the objective function \cite{ishikawa2009}.

\section{Non-linear Least Squares}
A robustified nonlinear least squares problem is the minimization of a function of the form\cite{ceres}:
\begin{equation}
  \label{eq:nlls}
  \sum_{i=1}^N \rho_i\left( ||f_i(\x_{P_i})||^2 \right).
\end{equation}
Here, $\rho_i$ is a {\em loss function}. If it is the identity function, the problem is an ordinary non-linear least squares problem. Otherwise, under some mild conditions on $\rho_i$,~\eqref{eq:nlls} can still be solved using a non-linear least squares algorithm after appropriate modifications to the residual vector and the Jacobian matrix~\cite{triggs2000}.

Observe that the first sum in ~\eqref{eq:problem} can be written in the form of~\eqref{eq:nlls} by setting
\begin{equation}
  f_i(x) = \frac{x - u_i}{\sqrt{2}\sigma} \qquad \text{and} \qquad \rho_i(s) = s,
\end{equation}
and the second sum by setting
\begin{equation}
  f_{k}(\x_P) = \b_k\T\x_P  \quad \text{and} \quad \rho_{k}(s) = \alpha_k\log\left(1 + \frac{s}{2} \right).
\end{equation}
Thus,~\eqref{eq:problem} is a robustified non-linear least squares problem.

\begin{table}
\centering
\begin{minipage}{250mm}
\hspace{-20mm}
\begin{tabular}{lcrrrrrrrr}
  \toprule
  \multicolumn{2}{c}{\multirow{2}{*}{\textbf{Method}}} & \multicolumn{2}{c}{\textbf{test001}} & \multicolumn{2}{c}{\textbf{test002}} & \multicolumn{2}{c}{\textbf{test003}} & \multicolumn{2}{c}{\textbf{test004}} \\
  & & obj.~~ & time~ & obj.~~ & time~  & obj.~~ & time~  & obj.~~ & time~ \\
  \midrule
  \multirow{2}{*}{
    \begin{minipage}{20mm}
      \raggedright
      As reported in~\cite{ishikawa2009}
    \end{minipage}
  } &                  
    \cite{ishikawa2011,ishikawa2009hocr}       
                              & 37769 & 1326 s.  & 25030 & 1330 s.  & 29805 & 1305 s. & 27356 & 1290 s.  \\
  & \cite{ishikawa2009}       & 38132 &   71 s.  & 24831 &   81 s.  & 29683 &   67 s. & 27354 &   79 s.  \\
  \midrule
  \multirow{3}{*}{
    \begin{minipage}{20mm}
      \raggedright
      On our computer
    \end{minipage}
  } &     
    \cite{ishikawa2011,ishikawa2009hocr}     
                            & 37691 &  625 s. & 24997 &  631 s. & 29762  & 623 s. & 27330 & 604 s.  \\
  & \cite{ishikawa2009}     & 37686 &  16 s.  & 25129 &  24 s.  & 29734  &  22 s. & 27219 &  18 s.  \\
  & Levenberg-Marquardt     & 37374 &   2 s.  & 24556 &   2 s.  & 29434  &   2 s. & 27088 &   2 s.  \\
  \bottomrule
\end{tabular}
\end{minipage}
\vspace{2mm}
\caption{Denoising the four 160${}\times{}$240 test images used in \cite{ishikawa2009hocr,ishikawa2009,ishikawa2011} (shown in Fig.~\ref{fig:example2x2}) with \foetwo\ filters, $K={}$3 and $\sigma={}$20. The table reports final objective function values and running times. 
}
\label{table:2x2}
\end{table}

\begin{figure*}
  \centering
  \def\imheight{36mm}
  \includegraphics[height=\imheight]{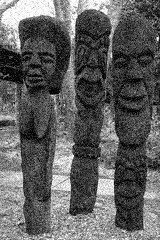}%
  \includegraphics[height=\imheight]{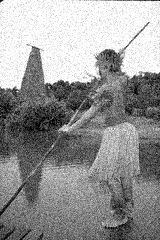}%
  \includegraphics[height=\imheight]{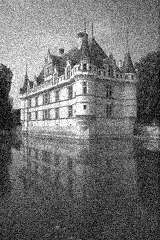}%
  \includegraphics[height=\imheight]{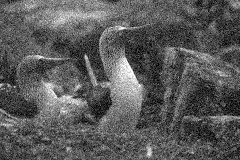}\\
  \includegraphics[height=\imheight]{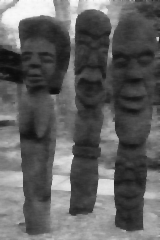}%
  \includegraphics[height=\imheight]{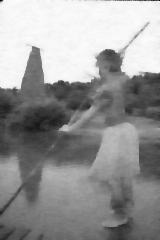}%
  \includegraphics[height=\imheight]{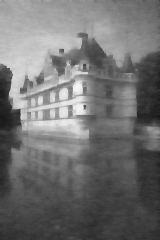}%
  \includegraphics[height=\imheight]{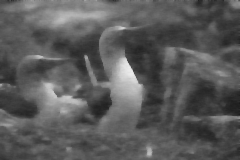}%
  \caption{The four denoised test images using a \foetwo\ FoE model. Table~\ref{table:2x2} shows the quantitative results.}
  \label{fig:example2x2}
\end{figure*}

\section{Experiments}

\subsection{Setup}
We used \textsc{ceres solver}~\cite{ceres} with the Levenberg-Marquardt algo\-rithm~\cite{nocedal2006} in combination with the \textsc{cholmod} sparse Cholesky factorization library~\cite{chen2008algorithm} to minimize the robustified non-linear least squares formulation of~\eqref{eq:problem}. 
We compare this to two state-of-the-art discrete optimization methods~\cite{ishikawa2009,ishikawa2011}.
The code for~\cite{ishikawa2011} is publicly available and we implemented~\cite{ishikawa2009} ourselves.

In our experiments, we used the noisy image as the initial point for the solver. Since the discrete methods use integer images, we rounded the continuous solution for the \foetwo\ filters to the nearest integer in $\{\text{0},\ldots,\text{255}\}$ to get a fair comparison.
This increased the final objective function value a little bit ($\!\!{}<{}$0.2\%).
All of the experiments were performed on a 3.47\thinspace GHz Intel Xeon and did not use any multi-threaded capabilities.

\begin{table}
\centering
\begin{minipage}{250mm}
\hspace{-15mm}
\begin{tabular}{lrrrrrrrr}
  \toprule
  \multirow{2}{*}{\textbf{\hspace{21mm}Method}} & \multicolumn{2}{c}{\textbf{test001}} & \multicolumn{2}{c}{\textbf{test002}} & \multicolumn{2}{c}{\textbf{test003}} & \multicolumn{2}{c}{\textbf{test004}} \\
  & obj.~~ & time~ & obj.~~ & time~  & obj.~~ & time~  & obj.~~ & time~ \\
  \midrule
  Levenberg-Marquardt, \foethree, $K=\text{8}$
                                        & 55186  &  25 s. & 39750  &  24 s. & 44798 &  25 s. & 42367 &  20 s. \\
  Levenberg-Marquardt, \foefive, $K=\text{24}$
                                        & 61304  & 132 s. & 42518  & 139 s. & 47093 & 149 s. & 44820 & 113 s. \\
  \bottomrule
\end{tabular}
\end{minipage}
\vspace{2mm}
\caption{Denoising with higher-order FoE models using the noisy images in Fig.~\ref{fig:example2x2}. 
The objective function values are from different optimization problems and are therefore not comparable.}
\label{table:3x3-5x5}
\end{table}

\begin{figure}
  \centering
  \def\svgwidth{0.5\linewidth}%
  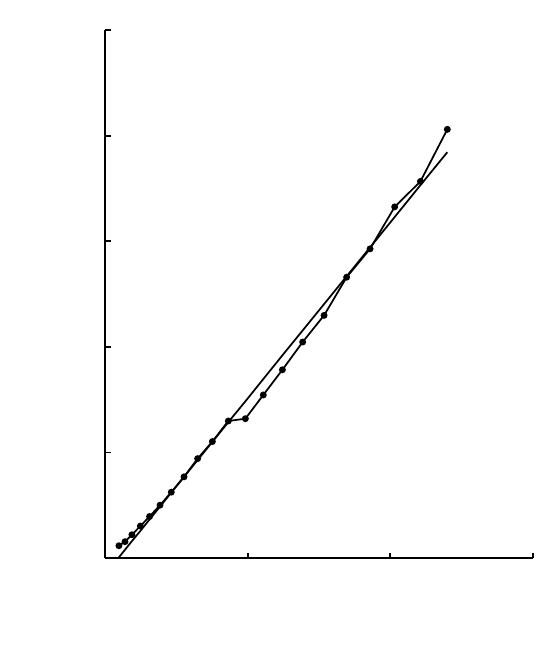%
  \def\svgwidth{0.5\linewidth}%
  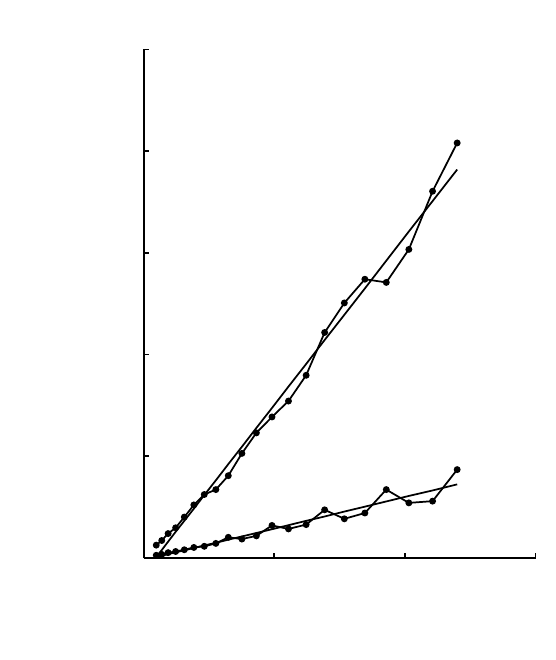%
  \caption{The time required to denoise an image is approximately linear in the number of pixels. The figure shows graphs for \foetwo~(left), \foethree\ and \foefive~(right) filters.}
  \label{fig:time}
\end{figure}

\subsection{Results}
We begin by denoising four test images using the \foetwo\ FoE model that has been commonly used for evaluating discrete methods. Table~\ref{table:2x2} shows the results, both as reported in~\cite{ishikawa2009} and of our own experiments and Fig.~\ref{fig:example2x2} illustrates the minima we found.
The  non-linear least squares solver finds a lower objective function value at a fraction of the runtime of the discrete methods.

Apart from the four test images used for benchmarking in previous publications, we added noise to the 100 test images in the Berkeley Segmentation Database \cite{berkeley}.
The non-linear least squares solver always found a better objective value than the method in \cite{ishikawa2009} (1\% on average) and was much faster (11 and~100 seconds, respectively, on average).

Because of limitations of the methods used for pseudo-boolean optimization, discrete methods for FoE inference have focused on the \foetwo\ case.
The methods in~\cite{fix2011},~\cite{ishikawa2011} and~\cite{kahl2012} are not capable of handling the \mbox{degree-9} polynomials that would be required for \foethree\ inference. 
In contrast, nothing is preventing a nonlinear least squares solver from using \foethree\ or \foefive\ filters. Table~\ref{table:3x3-5x5} contains our running times and final objective function values for the four test images.

Finally, we performed an experiment where we resized an image to different sizes, added noise, and denoised it using the nonlinear solver. Figure~\ref{fig:time} shows that the running time is approximately linear in the number of pixels.

\section{Discussion}
Non-linear least squares performs very well when applied to MAP Fields of Experts denoising. It is several times faster than the fastest method based on discrete optimization and it is immediately applicable to problems with larger filter sizes.

As pointed out by~\cite[Fig.~4]{kahl2012}, the more efficient reductions in~\cite{fix2011} do improve the speed of \cite{ishikawa2011}, but we observed at most a 1--2 second improvement when we used them with~\cite{ishikawa2009}.  The method presented in~\cite{kahl2012} also solves the individual pseudo-boolean problems better, but is slower.
Two other approaches exist, but they are both significantly slower~\cite{lan2006,potetz2007}.

While FoE denoising is a useful benchmark problem for discrete optimization, we should keep in mind that continuous methods can solve these problems much more efficiently.


\bibliographystyle{plain}
\bibliography{optimization}

\end{document}